# Ontology and Formal Semantics (*Integration Overdue*)†

WALID S. SABA

*American Institutes for Research*
wsaba@air.org

In this note we suggest that difficulties encountered in natural language semantics are, for the most part, due to the use of mere symbol manipulation systems that are devoid of any content. In such systems, where there is hardly any link with our commonsense view of the world, and it is quite difficult to envision how one can formally account for the considerable amount of content that is often implicit, but almost never explicitly stated in our everyday discourse.

The solution, in our opinion, is a compositional semantics grounded in an ontology that reflects *our commonsense view of the world and the way we talk about it in ordinary language*. In the compositional logic we envision there are ontological (or first-intension) concepts, and logical (or second-intension) concepts, and where the ontological concepts include not only Davidsonian events, but other abstract objects as well (e.g., states, processes, properties, activities, attributes, etc.)

It will be demonstrated here that in such a framework, a number of challenges in the semantics of natural language (e.g., metonymy, intensionality, metaphor, etc.) can be properly and uniformly addressed.

† **Draft – an extended version of** Saba, W. S. (2007), Compositional Semantics Grounded in Commonsense Metaphysics, 13th Portuguese Conference on Artificial Intelligence – EPIA 2007.

# 1 Introduction

Apparently, and perhaps for computational effectiveness, as Givon (1984) once suggested, in using ordinary spoken language to express our thoughts we tend to do so by using the least possible effort; by, for one thing, uttering the least number of words that are needed to convey a particular thought. Thus, for example, we make statements such as the following:

(1) a. *Simon is a rock.*[1]
b. *The ham sandwich wants another beer.*
c. *Sheba is articulate.*
d. *Jon bought a brick house.*
e. *Carlos likes to play bridge.*
f. *Jon visited a house on every street.*

Speakers of ordinary language, however, understand these sentences to mean the following, respectively:

(2) a. *Simon is* [as solid as] *a rock.*
b. *The* [person eating the] *ham sandwich wants another beer.*
c. *Sheba is* [an] *articulate* [person].
d. *Jon bought a house* [made of] *brick.*
e. *Carlos likes to play* [the card game] *bridge.*
f. *Jon visited a* [different] *house on every street.*

Clearly, any compositional semantics must somehow account for this [missing text], as such sentences are quite common and are not at all exotic, farfetched, or contrived. Traditionally, linguists and semanticists have dealt with such sentences by investigating various phenomena such as *metaphor* (2a); *metonymy* (2b); *textual entailment* (2c); *nominal compounds* (2d); *lexical ambiguity* (2e), and *quantifier scope ambiguity* (2f), to name a few. However, and although they seem to have a common denominator, it is somewhat surprising that in looking at the literature one finds that these phenomena have been studied quite independently; to the point where there is very little, if any, that seems to be common between the various proposals that are often suggested.

[1] 'Simon' as in 'Simon & Garfunkel'!

In our opinion this state of affairs is very problematic, as the prospect of a distinct paradigm for every single phenomenon in natural language cannot be realistically contemplated. Moreover, and as we hope to demonstrate in these notes, we believe that there is indeed a common symptom underlying these (and other) challenging problems in the semantics of natural language.

Before we make our case, let us at this very early juncture suggest this informal explanation for the missing text in (2): SOLID is (one of) the most salient features of a `rock` (2a); people, and not sandwiches, have 'wants', and EAT is the most salient relation that holds between a `person` and a `sandwich` (2b)[2]; `person` is the type of object of which ARTICULATE is the most salient property (2c); MADE-OF is the most salient relation between an `artifact` (house) and a `substance` (brick) (2d); PLAY is the most salient relation that holds between a `person` and a `game`, and not some `structure` (and, bridge is a `game`); and, finally, in the (possible) world that we live in, a `house` cannot be located on more than one `street`.

The point of this informal explanation is to suggest that the problem underlying most challenges in the semantics of natural language seems to lie in semantic formalisms that employ logics that are mere abstract symbol manipulation systems; systems that are devoid of any ontological content. What we suggest, instead, is a compositional semantics that is grounded in commonsense metaphysics, a semantics that views "logic as a language"; that is, a logic that has content, and ontological content, in particular, as has been recently and quite convincingly advocated by Cocchiarella (2001).

In these working notes we propose exactly such an approach. In particular, we will: (*i*) propose a semantics that is grounded in an ontology that reflects our commonsense view of reality and *the way we talk about it in ordinary language*; (*ii*) we will subsequently formalize the notion of 'salient property' and 'salient relation' and suggest how a strongly-typed compositional system can possibly utilize such information to explain some complex phenomena in natural language; (*iii*) it will be demonstrated that such a strongly-typed and ontologically grounded compositional semantics might also provide an explanation for the interpretation of anaphora without the need for complex discourse structures that may compromise compositionality; and (*iv*) we shall finally discuss how such an assumed ontological structure might be discovered, rather than invented, using natural language itself as a guide.

[2] Surely, a person can very much BUY, SELL, MAKE, PREPARE, WATCH, or HOLD, etc. a sandwich. Why EAT might be a more salient relation between a `person` and a `sandwich` is a question we shall pay considerable attention to below.

Before we commence, we need to make a few important notes. First, this is, to a large extent, 'work in progress', and thus it should not in any way be considered as a complete proposal. We are in fact putting out these ideas in the hope of soliciting some interest, and thus some novel ideas as to how this proposal might become a working theory. Second, as most of the ideas presented here are new, we chose to introduce our formalism in steps, through examples, modifying and extending our definitions as we proceed. It is hoped, however, that by the end we do achieve a somewhat coherent picture of a new compositional semantics that integrates commonsense reasoning in the interpretation of ordinary language.

# 2 Logical and Ontological Concepts

## 2.1 Types vs. Predicates

In our representation concepts belong to two quite distinct categories: (*i*) ontological concepts, such as `animal`, `substance`, `entity`, `artifact`, `event`, `state`, etc., which are assumed to exist in a subsumption hierarchy, and where the fact that an object of type `human` is (ultimately) an object of type `entity` is expressed as $\texttt{human} \sqsubseteq \texttt{entity}$; and (*ii*) logical concepts, which are the properties (that can be said) of and the relations (that can hold) between ontological concepts.

The following are examples that illustrate the difference between logical and ontological concepts, where the superscript `a` denotes abstract (or conceptual) existence:

(3) a. $\text{ARTICULATE}(x :: \texttt{human})$
b. $\text{ATTEND}(x :: \texttt{human}, y :: \texttt{event})$
c. $\text{IMMINENT}(x :: \texttt{event}^{\texttt{a}})$
d. $\text{DANCER}(x :: \texttt{human})$

This is what is meant by 'embedding' commonsense metaphysics into our semantics, as these predicates are supposed to reflect the fact that in our everyday discourse: the property ARTICULATE is ordinarily said of objects that must be of type **human** (3a); that we can always speak of objects of type **human** that attended some **event** (3b); that IMMINENT is ordinarily said of an **event** that does not (yet) actually exist; and DANCER is a property that is ordinarily said of objects that must be of type **human** (3d).

In addition to logical and ontological concepts, there are also proper nouns, which are the names of objects; objects that could be of any type. A proper noun, such as *sheba*, is interpreted as

(4) $[\![sheba]\!] \Rightarrow \lambda P[(\exists^1 x)(\text{NOO}(x :: \texttt{thing}, \text{`}sheba\text{'}) \wedge P(x :: \texttt{t}))]$,

where $\text{NOO}(x :: \texttt{thing}, s)$ is true of some individual object $x$ (which could be any **thing**), and $s$ if (the label) $s$ is the name of $x$, and **t** is presumably the type of objects that $P$ applies to (to simplify notation we will often write $[\![sheba]\!] \Rightarrow \lambda P[(\exists^1 sheba :: \texttt{thing})(P(sheba :: \texttt{t}))]$). Consider now the following, where we assumed $\text{THIEF}(x :: \texttt{human})$, i.e., that THIEF is a property that is ordinarily said of objects that must be of type **human**, and where $\mathbf{be}(x,y)$ is true when $x$ and $y$ are the same objects[3]:

(5) $[\![sheba\ is\ a\ thief]\!]$
$\Rightarrow (\exists^1 sheba :: \texttt{thing})(\exists x)(\text{THIEF}(x :: \texttt{human}) \wedge \mathbf{be}(sheba, x))$

This states that there is a unique object named *sheba*, some $y$ which must be an object of type **human**, such that $y$ is a THIEF and *sheba* is that $y$. Since $\text{EQ}(sheba, y)$, we could replace $y$ by the constant *sheba* obtaining the following:

(6) $[\![sheba\ is\ a\ thief]\!]$
$\Rightarrow (\exists^1 sheba :: \texttt{thing})(\exists x)(\text{THIEF}(x :: \texttt{human}) \wedge \mathbf{be}(sheba, x))$
$\Rightarrow (\exists^1 sheba :: \texttt{thing})(\text{THIEF}(sheba :: \texttt{human}))$

Note now that *sheba* is associated with more than one type in a single scope. In these situations a type unification must occur, where a type unification $(\texttt{s} \bullet \texttt{t})$ between two types $\texttt{s}$ and $\texttt{t}$, and where $Q \in \{\exists, \forall\}$, is defined (for now) as follows

(7)
$$
\begin{aligned}
&(Qx :: (\texttt{s} \bullet \texttt{t}))(P(x)) \\
&\equiv \begin{cases}
(Qx :: \texttt{s})(P(x)), & \textit{if } (\texttt{s} \sqsubseteq \texttt{t}) \\
(Qx :: \texttt{t})(P(x)), & \textit{if } (\texttt{t} \sqsubseteq \texttt{s}) \\
(Qx :: \texttt{s})(Qy :: \texttt{t})(\mathbf{R}(x,y) \wedge P(y)), & \textit{if } (\exists \mathbf{R})(\mathbf{R} = \textit{msr}(\texttt{s},\texttt{t})) \\
\bot, & \textit{otherwise}
\end{cases}
\end{aligned}
$$

where $\mathbf{R}$ is some salient relation that might exist between objects of type $\texttt{s}$ and objects of type $\texttt{t}$. That is, in situations where there is no subsumption relation between $\texttt{s}$ and $\texttt{t}$ the type unification results in keeping the variables of both types and in introducing some salient relation between them (we shall discuss these situations below). Finally when the type unification involves concrete (or actual) and abstract (or conceptual) existence the following rules apply:

(8)
$$
\begin{aligned}
&(\texttt{s} \bullet \texttt{t}^{\texttt{a}}) = (\texttt{s}^{\texttt{a}} \bullet \texttt{t}) = (\texttt{s} \bullet \texttt{t}) \\
&(\texttt{s}^{\texttt{a}} \bullet \texttt{t}^{\texttt{a}}) = (\texttt{s} \bullet \texttt{t})^{\texttt{a}}
\end{aligned}
$$

[3] We are using the fact that, when $a$ is a constant and $P$ is a predicate, $Pa \equiv \exists x[Px \wedge (x = a)]$ (see Gaskin, 1995).

Going to back to (6), the type unification in this case is actually quite simple, since $(\texttt{human} \sqsubseteq \texttt{thing})$:

(9)
$$
\begin{aligned}
&[\![\textit{sheba is a thief}]\!] \\
&\Rightarrow (\exists^1 \textit{sheba} :: \texttt{thing})(\text{THIEF}(\textit{sheba} :: \texttt{human})) \\
&\Rightarrow (\exists^1 \textit{sheba} :: (\texttt{thing} \bullet \texttt{human}))(\text{THIEF}(\textit{sheba})) \\
&\Rightarrow (\exists^1 \textit{sheba} :: \texttt{human})(\text{THIEF}(\textit{sheba}))
\end{aligned}
$$

In the final analysis, therefore, $\text{THIEF}(\textit{sheba})$, which is the standard first-order logic translation of 'Sheba is a thief' can be seen as an oversimplified representation of 'there is a unique object named *sheba*, an object that must be of type human, and such that *sheba* is a thief'. As an aside, since $\mathbf{be}(y, \textit{sheba})$, we could have also replaced *sheba* by $y$ in (5) resulting in $[\![\textit{sheba is a thief}]\!] \Rightarrow (\exists^1 y :: \texttt{human})(\text{THIEF}(y))$, which is an existential generalization of (9). That is, we would have obtained a weaker reading, namely that 'some object of type human (or 'someone') is a thief', which is an acceptable entailment of 'sheba is a thief'.

Finally, note the clear distinction between ontological concepts (such as `human`), which Cocchiarella (2001) calls first-intension concepts, and logical (or second-intension) concepts, such as $\textsc{thief}(x)$.

That is, what ontologically exist are objects of type `human`, not thieves, and THIEF is a mere property that we have come to use to talk of objects of type `human`[4]. Moreover, logical concepts such as THIEF are assumed to be defined by virtue of some logical expression, such as

$$(\forall x :: \texttt{human})(\textsc{thief}(x) \equiv_{df} \varphi),$$

where the exact nature of $\varphi$ might very well be susceptible to temporal, cultural, and other contextual factors, depending on what, at a certain point in time, a certain community considers an THIEF to be. In other words, while the property of being a THIEF that $x$ may exhibit is accidental (as well as temporal, cultural-dependent, etc.), the fact that some $x$ is an object of type `human` (and thus an `animal`, etc.) is not.

# 3 Semantics with Ontological Content

With the simple machinery introduced thus far we can actually start looking at how the embedding of ontological types in logical concepts coupled with simple type unifications can explain a number of phenomena that have proved problematic, such as intensionality, metonymy and reference resolution.

## 3.1 Intensional Verbs

Consider the following sentences and their corresponding translation into standard first-order logic:

(10) a. $[\![\textit{john found a unicorn}]\!] \Rightarrow (\exists x)(\textsc{unicorn}(x) \wedge \textsc{find}(jon, x))$
b. $[\![\textit{john sought a unicorn}]\!] \Rightarrow (\exists x)(\textsc{unicorn}(x) \wedge \textsc{seek}(jon, x))$

[4] Not recognizing the difference between logical (e.g., THIEF) and ontological concepts (e.g., `human`) is perhaps the reason why ontologies in most AI systems are rampant with multiple inheritance.

Note that $(\exists x)(\textsc{unicorn}(x))$ can be inferred in both cases, although it is clear that 'john sought a unicorn' should not entail the existence of a unicorn. In addressing this problem, Montague (1960) suggested treating *seek* as an intensional verb that more or less has the meaning of 'tries to find'; i.e. a verb of type $\langle\langle\langle e,t\rangle,t\rangle,\langle e,t\rangle\rangle$, using the tools of a higher-order intensional logic. To handle contexts where there are intensional as well as extensional verbs, mechanisms such as the 'type lifting' operation of Partee and Rooth (1983) were also introduced. The type lifting operation essentially coerces the types into the lowest possible type, the assumption being that if 'john sought *and* found' a unicorn, then a unicorn that was initially sought, but subsequently found, must have an existence.

In addition to unnecessary complication of the logical form, we believe the same intuition behind the 'type lifting' operation can be captured without the a priori separation of verbs into intensional and extensional ones, and in particular since most verbs seem to function intensionally and extensionally depending on the context. To illustrate this point further consider the following, where it is assumed that $\textsc{paint}(x :: \texttt{human}, y :: \texttt{entity}^{\texttt{a}})$; that is, it is assumed that the object of PAINT *does not necessarily* (*although it might*) *exist*:

(11) $[\![\textit{jon painted a dog}]\!]$

$$\begin{aligned}
&\Rightarrow (\exists^1 jon :: \texttt{human})(\exists d :: \texttt{dog}^{\texttt{a}})(\textsc{paint}(jon :: \texttt{human}, d :: \texttt{entity}^{\texttt{a}}))\\
&\Rightarrow (\exists^1 jon :: \texttt{human})(\exists d :: (\texttt{dog}^{\texttt{a}} \bullet \texttt{entity}^{\texttt{a}}))(\textsc{paint}(jon, d))\\
&\Rightarrow (\exists^1 jon :: \texttt{human})(\exists d :: (\texttt{dog} \bullet \texttt{entity})^{\texttt{a}})(\textsc{paint}(jon, d))\\
&\Rightarrow (\exists^1 jon :: \texttt{human})(\exists d :: \texttt{dog}^{\texttt{a}})(\textsc{paint}(jon, d))
\end{aligned}$$

Thus, 'Jon painted a dog' does not entail the existence of an actual dog, but a dog that only conceptually exists. However, consider now the following:

(12) $[\![\textit{jon painted his own dog}]\!]$

$$\begin{aligned}
&\Rightarrow (\exists^1 jon :: \texttt{human})(\exists d :: \texttt{dog}^{\texttt{a}})(\textsc{own}(jon :: \texttt{human}, d :: \texttt{entity}))\\
&\qquad \wedge \textsc{paint}(jon :: \texttt{human}, d :: \texttt{entity}^{\texttt{a}}))
\end{aligned}$$

Note that the types of PAINT and OWN are the same, except that the object of OWN is an object that must exist, while that of PAINT is an object that need not necessarily exist. The trivial type unification between these predicates (recall 8 above) would result in the following:

(13) $[\![ \textit{jon painted his own dog} ]\!]$
$\Rightarrow (\exists^1 jon :: \texttt{human})(\exists d :: \texttt{dog}^{\text{a}})$
$(\text{OWN}(jon :: \texttt{human}, d :: \texttt{entity}) \wedge \text{PAINT}(jon, d))$

Since $(\texttt{dog}^{\text{a}} \bullet \texttt{entity}) = \texttt{dog}$, the final type unification would result in the following:

(14) $[\![ \textit{jon painted his own dog} ]\!]$
$\Rightarrow (\exists^1 jon :: \texttt{human})(\exists d :: \texttt{dog})(\text{OWN}(jon, d) \wedge \text{PAINT}(jon, d))$

In the final analysis, therefore, Jon's paining of a dog does not entail the existence of a dog, but Jon's paining of his dog does; since Jon's owning of a dog entails its existence.

The point of the above example was to illustrate that the notion of intensional verbs can be captured in this simple formalism without the type lifting operation, particularly since an extensional interpretation might at times be implied even if the 'intensional' verb does not coexist with an extensional verb in the same context. To illustrate, let us assume $\text{PLAN}(x :: \texttt{human}, y :: \texttt{event}^{\text{a}})$; that is, that it always makes sense to say that some `human` is planning (or did plan) an `event` that need not (yet) actually exist. Consider now the following,

(15) $[\![ \textit{jon planned the trip} ]\!]$
$\Rightarrow (\exists^1 jon :: \texttt{entity})(\exists^1 e :: \texttt{trip}^{\text{a}})(\text{PLAN}(jon :: \texttt{human}, e :: \texttt{event}^{\text{a}}))$
$\Rightarrow (\exists^1 jon :: \texttt{human})(\exists^1 e :: (\texttt{trip}^{\text{a}} \bullet \texttt{event}^{\text{a}}))(\text{PLAN}(jon, e))$
$\Rightarrow (\exists^1 jon :: \texttt{human})(\exists^1 e :: \texttt{trip}^{\text{a}})(\text{PLAN}(jon, e))$

That is, saying 'john planned the trip' is simply saying that a specific object that must be a human has planned a specific trip, a trip that might not have actually happened[5]. Assuming $\text{LENGTHY}(e :: \texttt{event})$, however, i.e., that LENGTHY is a property that is ordinarily said of an (existing) event, then the interpretation of 'john planned the lengthy trip' should proceed as follows:

(16) $[\![ \textit{jon planned the lengthy trip} ]\!]$
$\Rightarrow (\exists^1 jon :: \texttt{entity})(\exists^1 e :: \texttt{trip}^{\text{a}})$

[5] Note that it is the trip (`event`) that did not necessarily happen, not the planning (`activity`) for it.

$$(\text{PLAN}(jon :: \mathtt{human}, e :: \mathtt{event}^{a}) \wedge \text{LENGTHY}(e :: \mathtt{event}))$$

Since $(\mathtt{trip}^{a} \bullet (\mathtt{event} \bullet \mathtt{event}^{a})) = (\mathtt{trip}^{a} \bullet \mathtt{event}) = \mathtt{trip}$ we finally get the following:

(17) $[\![\textit{jon planned the trip}]\!]$
$\Rightarrow (\exists^{1} jon :: \mathtt{human})(\exists^{1} e :: \mathtt{trip})(\text{PLAN}(jon, e) \wedge \text{LENGTHY}(e))$

That is, there is a specific human named *jon* that has planned a specific trip, a trip that was LENGTHY.

Finally, it should be noted here that the trip in (17) was finally considered to be an existing event due to other information contained in the same sentence. In general, however, this information can be contained in a larger discourse. For example, in interpreting 'John planned the trip. It was lengthy' the resolution of 'it' would force a retraction of the types inferred in processing 'John planned the trip', as the information that follows will 'bring down' the aforementioned `trip` from abstract to actual existence. This subject is clearly beyond the scope of this paper, but readers interested in the computational details of such processes are referred to (van Deemter & Peters, 1996).

### 3.2 On Dot (•) Objects

In addition to handling so-called intensional verbs, our proposal seems to also appropriately handle other situations that, on the surface, seem to be addressing a different issue. For example, consider the following:

(15) *jon read the book and then he burned it*

In Asher and Pustejovsky (2005) it is argued that this is an example of what they term *copredication*; which is the possibility of incompatible predicates to be applied to the same type of object. It is argued that in (15), for example, 'book' must have what is called a dot type, which is a complex structure that in a sense carries the 'informational content' sense (which is referenced when it is being read) as well as the 'physical object' sense (which is referenced when it is being burned). Elaborate machinery is then introduced to 'pick out' the right sense in the right context, and all in a well-typed compositional logic. But this approach presupposes that one can enumerate, *a priori*, all possible uses of the word 'book' in ordinary language[6]. Moreover, *copredication* seems to be a special case of metonymy, where the possible relations that could be implied are in fact much more constrained. An approach that can explain both notions, and hopefully without introducing much complexity into the logical form, should then be more desirable.

Let us first suggest the following:

(16) a. $\text{READ}(x :: \texttt{human}, y :: \texttt{content})$
b. $\text{BURN}(x :: \texttt{human}, y :: \texttt{physical})$

That is, we are assuming here that speakers of ordinary language understand 'read' and 'burn' as follows: it always makes sense to speak of a `human` that read some `content`, and of a human that burned some physical object. Consider now the following:

(16) $[\![\textit{jon read a book and then he burned it}]\!]$
$\Rightarrow (\exists^1 jon :: \texttt{entity})(\exists b :: \texttt{book})$
$(\text{READ}(jon :: \texttt{human}, b :: \texttt{content}))$
$\wedge\ \text{BURN}(jon :: \texttt{human}, b :: \texttt{physical}))$

The type unification of *jon* is straightforward, as the agent of BURN and READ are of the same type. Concerning b, however, there are a pair of type unifications $((\texttt{book} \bullet \texttt{physical}) \bullet \texttt{content})$ that must occur, that would result in the following:

(17) $[\![\textit{jon read a book and then he burned it}]\!]$
$\Rightarrow (\exists^1 jon :: \texttt{entity})(\exists b :: (\texttt{book} \bullet \texttt{content}))$
$(\text{READ}(jon, b) \wedge \text{BURN}(jon, b)))$

Again, since no subsumption relation exists between `book` and `content`, the two variables are kept and a salient relation between them is introduced, resulting in the following:

(18) $[\![\textit{jon read a book and then he burned it}]\!]$
$\Rightarrow (\exists^1 jon :: \texttt{entity})(\exists b :: \texttt{book})(\exists c :: \texttt{content})$

---

[6] Similar presuppositions are also made in a hybrid (connectionist/symbolic) 'sense modulation' approach described in (Rais-Ghasem & Corriveau, 1998).

$$(\mathbf{R}(b,c) \wedge \textsc{read}(jon,c) \wedge \textsc{burn}(jon,b))$$

That is, there is some unique object of type `human` (named *jon*), some book *b*, some content *c*, such that *c* is the content of *b*, and such that *jon* read *c* and burned *b*.

### 3.3 Reference Resolution and the Retraction of Type Inferences

Consider the following, where it is assumed that OWN is a relation that holds between objects of type `human` and objects of type `entity`:

(16) $[\![$*Jon owns Das Kapital*$]\!]$

$$\Rightarrow (\exists^1 jon :: \texttt{human})(\exists^1 dasKapital :: \texttt{book})$$
$$(\textsc{own}(jon :: \texttt{human}, dasKapital :: \texttt{entity}))$$
$$\Rightarrow (\exists^1 jon :: \texttt{human})(\exists^1 dasKapital :: (\texttt{book} \bullet \texttt{entity}))$$
$$(\textsc{own}(jon, dasKapital))$$
$$\Rightarrow (\exists^1 jon :: \texttt{human})(\exists^1 dasKapital :: \texttt{book})(\textsc{own}(jon, dasKapital))$$

Quite simply, then, '*Jon owns Das Kapital*' is interpreted as follows: there is an object of type `human` named *jon*, and an object of type book named *dasKapital*, and *jon* owns *dasKapital*. However, consider now the following:

(17) $[\![$*Jon owns Das Kapital but he does not agree with it.*$]\!]$

$$\Rightarrow (\exists^1 jon :: \texttt{human})(\exists^1 dasKapital :: \texttt{book})(\textsc{own}(jon, dasKapital))$$
$$\wedge (\exists^1 he :: \texttt{human})(\exists^1 it :: \texttt{entity})$$
$$(\neg\textsc{agree}(he :: \texttt{human}, it :: \texttt{content}))$$

Resolving 'he' with Jon is straightforward since the agent of AGREE must be an object of type `human`. Resolving 'it' with Das Kapital, however, is not straightforward. This is how things look when attempting to link the two sentences:

(18) $[\![$*Jon owns Das Kapital but he does not agree with it.*$]\!]$

$$\Rightarrow (\exists^1 jon :: \texttt{human})(\exists^1 dasKapital :: (\texttt{book} \bullet \texttt{content}))$$
$$(\textsc{own}(jon, dasKapital) \wedge \neg\textsc{agree}(jon, dasKapital))$$

Again, since there is no subsumption relation between (the physical object) book and (the abstract object) content, the type unification results in keeping the two variables, one for each type, and in introducing some salient relation between them:

(19) $[\![$*Jon owns Das Kapital but he does not agree with it.*$]\!]$
$\Rightarrow (\exists^1 jon :: \texttt{human})(\exists^1 dasKapital :: \texttt{book})(\exists^1 c :: \texttt{content})$
$(\mathbf{R}(dasKapital, c) \wedge \textsc{own}(jon, dasKapital) \wedge \neg \textsc{agree}(jon, c))$

In the final analysis, therefore, 'Jon owns Das Kapital but he does not agree with it' is interpreted as follows: Jon owns the physical book 'Das Kapital' and he does not agree with its content!

### 3.4 Metonymy

Consider the following, where $\textsc{want}(x :: \texttt{human}, y :: \texttt{thing}^{a})$, i.e., where it is assumed that WANT is a relation that holds between objects of type **human** and objects that could be any **thing**, and where the object of wanting does not necessarily exist:

(15) $[\![$*the ham sadnwich wants a beer*$]\!]$
$\Rightarrow (\exists^1 x :: \texttt{hamSandwich})(\exists^1 y :: \texttt{beer}^{a})(\textsc{want}(x :: \texttt{human}, y :: \texttt{thing}^{a}))$

Note now that $x$ and $y$ in (14) are associated with more than one type in a single scope, and thus a type unification must occur. Assuming $\texttt{beer} \sqsubseteq \texttt{thing}$ the type unification concerning $y$ is straightforward, although beer should now be considered an object that need not necessarily exist (as it is the object of a WANT):

$[\![$*the ham sadnwich wants a beer*$]\!]$
$\Rightarrow (\exists^1 x :: (\texttt{hamSandwich} \bullet \texttt{human}))(\exists^1 y :: (\texttt{beer} \bullet \texttt{thing}^{a}))(\textsc{want}(x, y))$
$\Rightarrow (\exists^1 x :: (\texttt{hamSandwich} \bullet \texttt{human}))(\exists^1 y :: \texttt{beer}^{a})(\textsc{want}(x, y))$

The type unification concerning $x$ is not, however, as simple, since there is no subsumption relationship between the types concerned. As usual, the type unification will result in keeping two variables, one for each type, and in introducing some relation between them:

$[\![$*the ham sadnwich wants a beer*$]\!]$

$\Rightarrow (\exists^1 x :: (\texttt{hamSandwich} \bullet \texttt{human}))(\exists^1 y :: \texttt{beer}^{\texttt{a}})(\text{WANT}(x, y))$
$\Rightarrow (\exists^1 x :: \texttt{hamSandwich})(\exists^1 z :: \texttt{human})(\exists^1 y :: \texttt{beer})\ (\mathbf{R}(x, z) \wedge \text{WANT}(z, y))$
where $\mathbf{R} = msr(\texttt{human}, \texttt{sandwich})$, i.e., where $\mathbf{R}$ is assumed to be some salient relation (e.g., EAT, ORDER, etc.) that exists between an object of type `human`, and an object of type `sandwich` (more on this below).

# 4 All Objects were Created Equal

## 4.1 Reference to Abstract Objects

In a recent argument *Against Fantology*, Smith (2005) notes that too much attention has been paid to the false doctrine that much can be discovered about the ontological structure of reality by predication in first-order logic. According to Smith, for example, the use of standard predication in first-order logic in the following

(20) a. $[\![\textit{sheba is a singer}]\!] \Rightarrow \text{SINGER}(\textit{sheba})$
b. $[\![\textit{sheba is a cat}]\!] \Rightarrow \text{CAT}(\textit{sheba})$
c. $[\![\textit{sheba is old}]\!] \Rightarrow \text{OLD}(\textit{sheba})$

completely masks the fact that in these sentences we are referring to: one of Sheba's activities, namely her singing (20a); the class in which Sheba is a member (20b); and to one of Sheba's attributes, namely her age (20c). To make this point more acute one must consider a more realistic discourse. As an example, consider the following:

(21) *Sheba is an old dancer.*
a. *That has not made her less popular, however.*
b. *She has been doing that for more than* 40 *years.*

We argue that *that* in (21a) must clearly be referring to an *attribute* of Sheba, namely her age, while *that* in (21b) must clearly be referring to Sheba's singing (*activity*). This suggests that the definition of the logical concepts OLD and DANCER must have some internal structure that somehow refers to the 'age' attribute and the 'dancing' activity, respectively. Moreover, and as argued by Larson (1995), a statement such as 'Sheba is an old dancer' is ambiguous in that could mean any of the following:

(22) *Sheba is an old dancer.*
$\Rightarrow$ *Sheba is a dancer and an old person*
$\Rightarrow$ *Sheba's dancing is old (she has been dancing for a long time)*

That is, 'old' seems to modify both, Sheba, and the dancing activity that Sheba performs, which again suggests quantifying over abstract objects such as attributes, activities, etc.

In this respect we not only agree with Davidson's (1980) suggestion that "there is a lot of language we can make sense of if we suppose events exist", we further suggest that there is even a lot more language we can make sense of if we also admit the existence of other abstract entities, such as attributes, activities, processes, states, etc. Thus, the interpretation

$$[\![\textit{sheba is an old dancer}]\!] \Rightarrow (\exists^1 sheba :: \texttt{human})(\text{OLD}(\text{DANCER}(sheba)))$$

should be thought of as a 'condensed form' that, when expanded, would contain the logical expressions corresponding to the definitions of the logical concepts OLD and DANCER, which could be defined as follows, where $\triangle$ could be thought of as a 'typicality' operator:

(23) a. $(\forall x :: \texttt{human})(\text{DANCER}(x)$
$\equiv_{df} (\exists a :: \texttt{activity})(\text{DANCING}(a) \wedge \triangle(\mathbf{do}(x,a)) \wedge \phi(x,a...)))$

b. $(\forall x :: \texttt{entity})(\text{OLD}(x)$
$\equiv (\exists a :: \texttt{attribute})(\text{AGE}(a) \wedge \lambda P[P(x) \wedge (\exists y :: \texttt{physical})$
$(P(y) \wedge \text{TYPICAL}_{\text{AGE}}(y) \wedge (x \circ \text{AGE} >> y \circ \text{AGE}))])))$

That is, some human $x$ is/was a dancer iff $x$ often does/did perform an `activity` we call DANCING (23a); and any `physical` object $x$ that is a $P$ is old iff there is some other `physical` object $y$ that is also a $P$ and where $\text{TYPICAL}_{\text{AGE}}(y)$ ; i.e., where $y$ is a typical object as far as the attribute AGE is concerned, such that the age of $x$ is considerably greater than that of $y$ (23b)[7].

---

[7] Our specific formulation of the logical concepts DANCER and OLD (as well as others) could be questioned without affecting our overall argument. This is in fact precisely the whole point, namely that, unlike ontological concepts, the exact definition of a logical concept might very well depend on cultural, temporal, and perhaps individual factors (individuals perhaps do differ in the formulation of concepts such as 'liberal', 'generous', 'educated', etc.) What should not be questionable is the fact that speaking of a SINGER, for example,

Assuming now that $s \Rightarrow \texttt{a}$ and $s \rightarrow \texttt{a}$ refer, respectively, to what we might call a 'condensed' and an 'expanded' interpretation of a sentence $s$ into the logical expression $\texttt{a}$, we suggest the following:

(24) a. $[\![\textit{liz is famous}]\!]$
$\Rightarrow (\exists^1 liz :: \texttt{human})(\textsc{famous}(liz))$
$\rightarrow (\exists^1 liz :: \texttt{human})(\exists^1 p :: \texttt{property})(\textsc{fame}(p) \wedge \mathbf{has}(liz, p))$

b. $[\![\textit{aging is inevitable}]\!]$
$\Rightarrow (\exists^1 x :: \texttt{process})(\textsc{aging}(x) \wedge \textsc{inevitabile}(x))$
$\rightarrow (\exists^1 x :: \texttt{process})(\exists^1 y :: \texttt{property})$
$(\textsc{aging}(x) \wedge \textsc{inevitability}(y) \wedge \mathbf{has}(x, y))$

c. $[\![\textit{jon is aging}]\!]$
$\Rightarrow (\exists^1 jon :: \texttt{human})(\textsc{aging}(jon))$
$\rightarrow (\exists^1 jon :: \texttt{human})(\exists^1 x :: \texttt{process})(\textsc{aging}(x) \wedge \mathbf{gt}(jon, x))$

d. $[\![\textit{fame is desirable}]\!]$
$\Rightarrow (\exists^1 x :: \texttt{property})(\textsc{fame}(x) \wedge \textsc{desirable}(x))$
$\rightarrow (\exists^1 x :: \texttt{property})(\exists^1 y :: \texttt{property})$
$(\textsc{fame}(x) \wedge \textsc{desirability}(y) \wedge \mathbf{has}(x, y))$

e. $[\![\textit{sheba is a dancer}]\!]$
$\Rightarrow (\exists^1 sheba :: \texttt{human})(\textsc{dancer}(sheba))$
$\rightarrow (\exists^1 sheba :: \texttt{human})(\exists^1 a :: \texttt{activity})$
$(\textsc{dancing}(a) \wedge \Delta(\mathbf{do}(sheba, a)))$

f. $[\![\textit{sheba is dead}]\!]$
$\Rightarrow (\exists^1 sheba :: \texttt{human})(\textsc{dead}(sheba))$
$\rightarrow (\exists^1 sheba :: \texttt{human})(\exists^1 s :: \texttt{physioState})$
$(\textsc{death}(s) \wedge \mathbf{in}(sheba, s))$

These interpretations can be explained as follows: some unique object of type `human`, named *liz*, has a certain `property`, namely FAME (13a); INEVITABILITY is a property that the `process` of AGING has (13b); a unique object of type `human`, named *jon*, is going through the process of AGING (13c); DESIRABILITY is a `property` that another `property`, namely FAME, has (13d); a unique object of type `human`, named *sheba*, often does perform the DANCING `activity` (13e); and a unique object of type `human`, named *sheba*, is in a certain state, namely DEATH.

must entail speaking of some `activity` that made an object a SINGER, as example (21) suggests.

Note now that a sentence such as '*Sheba is an old singer*' could be interpreted as follows:

(25) $[\![ \textit{sheba is an old dancer} ]\!]$
$\Rightarrow (\exists^1 sheba :: \texttt{human})(\textsc{old}(\textsc{dancer}(sheba)))$
$\Rightarrow (\exists^1 sheba :: \texttt{human})(\exists a :: \texttt{activity})$
$(\textsc{dancing}(a) \wedge \Delta(\mathbf{do}(sheba, a) \wedge (\textsc{old}(a) \vee \textsc{old}(sheba)))$

That is, OLD could predicate Sheba or her dancing `activity`, and where OLD is that defined in (23).

As another example that illustrates the need for quantifying over activities, consider the following example:

(26) *jon did not paint a dog*

We argue that (26) can be true if *jon* painted something other than a dog, or if *jon* did not paint anything at all. The interpretation of (26) should initially proceed as follows:

(27) $[\![ \textit{jon did not paint a dog} ]\!]$
$\Rightarrow (\exists^1 jon :: \texttt{human})(\exists d :: \texttt{dog}^{\mathsf{a}})(\neg\textsc{paint}(jon :: \texttt{human}, d :: \texttt{entity}^{\mathsf{a}}))$

That is, there is a unique object, *jon* (of type `human`), and some object $d$ of type `dog` (an object that need not necessarily exist), such that it is not true that *jon* painted $d$. Assuming that 'paint' is defined in terms of some activity as follows:

$$\textsc{paint}(x :: \texttt{human}, y :: \texttt{entity}^{\mathsf{a},n})$$
$$\equiv (\exists^n a :: \texttt{activity})(\textsc{painting}(a) \wedge \mathbf{do}(a, x) \wedge \mathbf{theme}(a, y))\ ,$$

Since $\neg(\exists x)(P(x) \wedge Q(x)) \equiv (\forall x)(\neg P(x) \vee \neg Q(x))$ we then get the following:

$$\neg\textsc{paint}(x :: \texttt{human}, y :: \texttt{entity}^{\mathsf{a}})$$
$$\equiv \neg((\exists a :: \texttt{activity})(\textsc{painting}(a) \wedge \mathbf{do}(a, x) \wedge \mathbf{theme}(a, y)))$$
$$\equiv (\forall a :: \texttt{activity})(\neg\textsc{painting}(a) \vee \neg\mathbf{do}(a, x) \vee \neg\mathbf{theme}(a, y))$$
$$\equiv (\forall a :: \texttt{activity})(\textsc{painting}(a) \supset \neg\mathbf{do}(a, x) \vee \neg\mathbf{theme}(a, y))$$

Thus, (27) can now be written as follows:

(28) $[\![\textit{jon did not paint a dog}]\!]$
$\Rightarrow (\exists^1 jon :: \texttt{human})(\exists d :: \texttt{dog}^{\texttt{a}})(\neg\text{PAINT}(jon :: \texttt{human}, d :: \texttt{entity}^{\texttt{a}}))$
$\Rightarrow (\exists^1 jon :: \texttt{human})(\exists d :: \texttt{dog}^{\texttt{a}})((\forall a :: \texttt{activity})$
$(\text{PAINTING}(a) \supset \neg\mathbf{do}(a, jon) \vee \neg\mathbf{theme}(a, d)))$

In the final analysis, therefore, '*jon did not paint a dog*' entails that either *jon* was not the agent of any painting activity, or the theme of the painting was not a dog, which is a plausible interpretation of (28).

## 4.2 Wise Activities? (Abstract Objects and Metonymy)

Consider the following:

(29) a. *exercising is wise*
b. *jon is exercising*

---

c. *jon is wise*

Clearly the above inference is valid, although one can hardly think of attributing the `property` WISE to an `activity` (exercising). Let us see how we might explain this argument. We start with the simplest:

(30) $[\![\textit{jon is exercising}]\!]$
$\Rightarrow (\exists^1 jon :: \texttt{human})(\exists^1 act :: \texttt{activity})$
$(\text{EXERCISING}(act) \wedge \text{AGENT}(act, jon))$

Let us now consider the following:

(31) $[\![\textit{exercising is wise}]\!]$
$\Rightarrow (\forall a :: \texttt{activity})(\text{EXERCISING}(a)$
$\supset (\exists^1 p :: \texttt{property})(\text{WISDOM}(p) \wedge \mathbf{has}(a :: \texttt{human}, p))$

That is, any exercising `activity` has a property, namely WISDOM, which is a property that ordinarily an object of type `human` has. Note, however, that a type unification for the variable *a* must now occur:

(32) $[\![\textit{exercising is wise}]\!]$
$\Rightarrow (\forall a :: (\texttt{human} \bullet \texttt{activity}))(\text{EXERCISING}(a)$
$\supset (\exists^1 p :: \texttt{property})(\text{WISDOM}(p) \wedge \mathbf{has}(a, p))$

The most salient relation between a $\texttt{human}$ and an $\texttt{activity}$ is that of agency: a human is typically the AGENT of an activity:

(33) $[\![$*exercising is wise*$]\!]$
$$\Rightarrow (\forall a :: \texttt{activity})(\forall x :: \texttt{human}^{\texttt{a}})(\text{EXERCISING}(a) \wedge \text{AGENT}(a, x)$$
$$\supset (\exists^1 p :: \texttt{property})(\text{WISDOM}(p) \wedge \mathbf{has}(x, p))$$

Essentially, therefore, we get the following: any human $x$ has the property of being WISE whenever $x$ is the agent of an exercising activity. Note now that (30), (33) and modes ponens results in the following, which is the meaning of '*jon is wise*':

$$(\exists^1 jon :: \texttt{human})(\exists^1 p :: \texttt{property})(\text{WISDOM}(p) \wedge \mathbf{has}(x, p))$$

Finally, note that the inference in (33) was proven valid only after uncovering the missing text, since '*exercising is wise*' was essentially interpreted as '[any human that performs the activity of] *exercising is wise*'.

### 4.3 90 *is not rising*

Consider the following erroneous entailment:

(34) a. *the temperature is* 90
b. *the temperature is rising*

---

c. 90 *is rising*

We argue that the source of this erroneous entailment is due to the fact that the copula 'is' is being used on different types of objects in (34a) and (35b). When $x$ and $y$ are of the same type (or one subsumes the other!) then 'is' in '$x$ is $y$' represents identity, otherwise some other relation that holds between the two objects, as the examples in (24) also illustrate. More to the point consider the following:

(35) $[\![$*the temperature is* 90$]\!]$
$$\Rightarrow (\exists^1 x :: \texttt{temperature})(\exists^1 y :: \texttt{measure})(\text{VALUE}(y, 90) \wedge \mathbf{be}(x, y))$$
$$\Rightarrow (\exists^1 x :: (\texttt{temperature} \bullet \texttt{measure}))(\text{VALUE}(x, 90))$$
$$\Rightarrow (\exists^1 x :: \texttt{temperature})(\text{VALUE}(x, 90))$$

That is, since ($\texttt{temperature} \sqsubseteq \texttt{measure}$), the variables $x$ and $y$ unify, laving one variable, and thus EQ is effectively reduced to identity. However, in 'the temperature is rising' we are saying no more than 'the temperature *is in the process of* rising':

(36) $[\![\textit{the temperature is rising}]\!]$
$\Rightarrow (\exists^1 x :: \texttt{temperature})(\exists^1 p :: \texttt{process})(\text{RISING}(p) \wedge \mathbf{gt}(x,y))$

# 5 Types and Salient Relations

Thus far we have assumed the existence of a function $\texttt{msr}(\texttt{s},\texttt{t})$ that returns, if it exists, the most salient relation $\mathbf{R}$ between two types $\texttt{s}$ and $\texttt{t}$. In this section we suggest what this function might look like.

Before we proceed, however, we need to extend the notion of assigning ontological types to properties and relations slightly. First, consider the following:

(37) a. *Pass that car, will you. He is really annoying me.*
b. *Pass that car, will you. They are really annoying me.*

We argue that 'he' in (34a) refers to 'the person *driving* [that] car' while 'they' in (34b) refers to 'the people *riding* in [that] car'. The question here is this: although there are many possible relations between a person and a car (e.g., DRIVE, RIDE, MANUFACTURE, DESIGN, MAKE, etc.) how is it that DRIVE is the one that most speakers assume in (37a), while RIDE is the one most speakers would assume in (37b)? Here's a plausible answer:

- DRIVE is more salient than RIDE, MANUFACTURE, DESIGN, MAKE, etc. since the other relations apply higher-up in the hierarchy; that is, the fact that we MAKE a `car`, for example, is not due to `car`, but to the fact that MAKE can be said of any `artifact` and ($\texttt{car} \sqsubseteq \texttt{artifact}$).

- While DRIVE is a more salient relation between a **human** and a **car** than RIDE, most speakers of ordinary English understand the DRIVE relation to hold between one human and one car (at a specific point in time), while RIDE is a relation that holds between many (several, or few!) people and one car. Thus, 'they' in (37b) fails to unify with DRIVE, and RIDE, which is the next most salient relation is then picked out.

In other words, the type assignments of DRIVE and RIDE are understood by speakers of ordinary language as follows:

$\text{DRIVE}(x :: \mathbf{human}^{1}, y :: \mathbf{car}^{1})$
$\text{RIDE}(x :: \mathbf{human}^{1+}, y :: \mathbf{car}^{1})$

With this background, let us now suggest how the function $\mathtt{msr}(\mathtt{s},\mathtt{t})$ that picks out the most salient relation **R** between two types $\mathtt{s}$ and $\mathtt{t}$ is computed.

We say $\mathtt{pap}(\text{P},\mathtt{t})$ when the property P applies to objects of type $\mathtt{t}$, and $\mathtt{rap}(\text{R},\mathtt{s},\mathtt{t})$ when the relation R holds between objects of type $\mathtt{s}$ and objects of type $\mathtt{t}$. We define a list $\mathtt{lpap}(\mathtt{t})$ of all properties that apply to objects of type $\mathtt{t}$, and $\mathtt{lrap}(\mathtt{s},\mathtt{t})$ of all relations that hold between objects of type $\mathtt{s}$ and objects of type $\mathtt{t}$, as follows:

(38) $\mathtt{lpap}(\mathtt{t}) = [\text{P} \mid \mathtt{pap}(\text{P},\mathtt{t})]$
$\mathtt{lrap}(\mathtt{s},\mathtt{t}) = [\langle \text{R}, m, n\rangle \mid \mathtt{rap}(\text{R},\mathtt{s}^{m},\mathtt{t}^{n})]$

The lists (of lists) $\mathtt{lpap}^{*}(\mathtt{t})$ and $\mathtt{lrap}^{*}(\mathtt{s},\mathtt{t})$ can now be inductively defined as follows:

(39) $\mathtt{lpap}^{*}(\mathtt{thing}) = [\ ]$
$\mathtt{lpap}^{*}(\mathtt{t}) = \mathtt{lpap}(\mathtt{t}) : \mathtt{lpap}^{*}(\mathtt{sup}(\mathtt{t}))$

$\mathtt{lrap}^{*}(\mathtt{s},\mathtt{thing}) = [\ ]$
$\mathtt{lrap}^{*}(\mathtt{s},\mathtt{t}) = \mathtt{lrap}(\mathtt{s},\mathtt{t}) : \mathtt{lrap}^{*}(\mathtt{s},\mathtt{sup}(\mathtt{t}))$

where $(e : s)$ is a list that results from attaching the object $e$ to the front of the (ordered) list $s$, and where $\mathtt{sup}(\mathtt{t})$ returns the immediate (and single!) parent of $\mathtt{t}$. Finally, we define the function $\mathtt{msr}(\langle \mathtt{s}^{m}, \mathtt{t}^{n}\rangle)$ which returns most the salient relation between objects of type $\mathtt{s}$ and $\mathtt{t}$, with constraints $m$ and $n$, respectively, as follows:

$$\texttt{msr}(\langle \texttt{s}^m, \texttt{t}^n \rangle) = \texttt{if}\ (\texttt{s} \neq [\ ])\ \texttt{then}\ (\texttt{head s})\ \texttt{else}\ \perp$$
$$\textit{where}$$
$$\texttt{s} = [\text{R} \mid \langle \text{R}, a, b\rangle \in \texttt{lrap}^*(\texttt{s},\texttt{t}) \wedge (a \geq m) \wedge (b \geq n)]$$

Assuming now the ontological and logical concepts shown in figure 1, for example, then

$$\texttt{lpap}^*(\texttt{human}) = [[\text{ARTICULATE},...],[\text{HUNGRY},...],[\text{HEAVY},...],,[\text{OLD},...],...]$$
$$\texttt{lrap}^*(\texttt{human},\texttt{car}) = [[\langle\text{DRIVE},1,1\rangle,...],[\langle\text{RIDE},1^+,1\rangle,...],,...]$$

These lists are ordered, and the degree to which a property or a relation is salient is inversely related to the position of the property or the relation in the list. Thus, for example, while a **human** may DRIVE, MAKE, BUY, SELL, BUILD, etc. a **car**, DRIVE is a more salient relation between a **human** and a **car** than RIDE, which, in turn, is more salient than MANUFACTURE, MAKE, etc. Moreover, assuming the above sets we have

$$\texttt{msr}(\langle \texttt{human}^1, \texttt{car}^1\rangle) = \text{DRIVE}$$
$$\texttt{msr}(\langle \texttt{human}^{1+}, \texttt{car}^1\rangle) = \text{RIDE}$$

which essentially says DRIVE is the most salient relation in a context where we are speaking of a single human and a single car, is that of

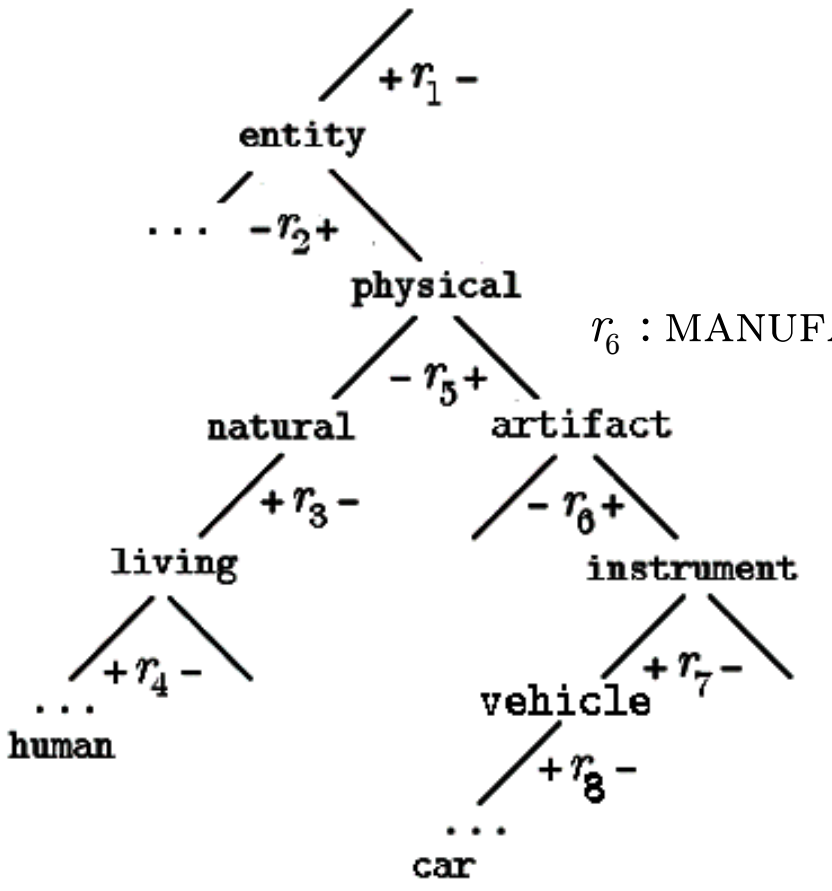

$r_1$ : OLD($x$ :: **entity**)
$r_2$ : HEAVY($x$ :: **physical**)
$r_3$ : HUNGRY($x$ :: **living**)
$r_4$ : ARTICULATE($x$ :: **human**)
$r_5$ : MAKE($x$ :: **human**, $y$ :: **artifact**)
$r_6$ : MANUFACTURE($x$ :: **human**, $y$ :: **instrument**)
$r_7$ : RIDE($x$ :: **human**, $y$ :: **vehicle**)
$r_8$ : DRIVE($x$ :: **human**, $y$ :: **car**)

**Figure 1.** Logical and ontological concepts.

Note, therefore, that 'They are annoying me' in (37b) is interpreted as follows:

$$
\begin{aligned}
&[\![\textit{They are annoying me}]\!] \\
&\Rightarrow (\exists \mathit{they} :: (\mathtt{human}^{1+} \bullet \mathtt{car}))(\exists \mathit{me} :: \mathtt{human}^{1})(\textsc{annoying}(\mathit{they}, \mathit{me})) \\
&\Rightarrow (\exists \mathit{they} :: \mathtt{human}^{1+})(\exists c :: \mathtt{car})(\exists \mathit{me} :: \mathtt{human}^{1}) \\
&\qquad\qquad (\textsc{riding}(\mathit{they}, c) \wedge \textsc{annoying}(\mathit{they}, \mathit{me}))
\end{aligned}
$$

# 6 Discussion

If the main business of semantics is to explain how linguistic constructs relate to the world, then semantic analysis of natural language text is, indirectly, an attempt at uncovering the semiotic ontology of commonsense knowledge, and particularly the background knowledge that seems to be implicit in all that we say in our everyday discourse. While this intimate relationship between language and the world is generally accepted, semantics (in all its paradigms) has traditionally proceeded in one direction: by first stipulating an assumed set of ontological commitments followed by some machinery that is supposed to, somehow, model meanings in terms of that stipulated structure of reality.

With the gross mismatch between the trivial ontological commitments of our semantic formalisms and the reality of the world these formalisms purport to represent, it is not surprising therefore that challenges in the semantics of natural language are rampant. However, as correctly observed by Hobbs (1985), semantics could become nearly trivial if it was grounded in an ontological structure that is "isomorphic to the way we talk about the world". The obvious question however is 'how does one arrive at this ontological structure that implicitly underlies all that we say in everyday discourse?' One plausible answer is the (seemingly circular) suggestion that the semantic analysis of natural language should itself be used to uncover this structure. In this regard we strongly agree with Dummett (1991) who states:

> We must not try to resolve the metaphysical questions first, and then construct a meaning-theory in light of the answers.

> We should investigate how our language actually functions, and how we can construct a workable systematic description of how it functions; the answers to those questions will then determine the answers to the metaphysical ones.

What this suggests, and correctly so, in our opinion, is that in our effort to understand the complex and intimate relationship between ordinary language and everyday commonsense knowledge, one could, as also suggested in (Bateman, 1995), "use language as a tool for uncovering the semiotic ontology of commonsense" since ordinary language is the best known theory we have of everyday knowledge.

To avoid this seeming circularity (in wanting this ontological structure that would trivialize semantics; while at the same time suggesting that semantic analysis should itself be used as a guide to uncovering this ontological structure), we suggested here performing semantic analysis from the ground up, assuming a minimal (almost a trivial and basic) ontology, in the hope of building up the ontology as we go guided by the results of the semantic analysis. The advantages of this approach are: (*i*) the ontology thus constructed as a result of this process would not be invented, as is the case in most approaches to ontology (*e.g.*, Lenat, & Guha (1990); Guarino (1995); and Sowa (1995)), but would instead be discovered from what is in fact implicitly assumed in our use of language in everyday discourse; (*ii*) the semantics of several natural language phenomena should as a result become trivial, since the semantic analysis was itself the source of the underlying knowledge structures (in a sense, the semantics would have been done before we even started!)

Throughout this paper we have tried to demonstrate that a number of challenges in the semantics of natural language can be easily tackled if semantics is grounded in a strongly-typed ontology that reflects our commonsense view of the world and the way we talk about it in ordinary language. Our ultimate goal, however, is the systematic *discovery* of this ontological structure, and, as also argued in Saba (2007), it is the systematic investigation of how ordinary language is used in everyday discourse that will help us discover (as opposed to invent) the ontological structure that seems to underlie all what we say in our everyday discourse.